\documentclass[runningheads,a4paper]{llncs}

\usepackage{amssymb}
\usepackage{graphicx}
\usepackage{subcaption}
\usepackage{placeins}
\usepackage{color}
\usepackage{fixltx2e}
\usepackage{url}
\usepackage{placeins}
\usepackage[table]{xcolor}
\usepackage{multirow}
\usepackage{booktabs}
\usepackage[hidelinks]{hyperref}
\usepackage{todonotes}
\usepackage[export]{adjustbox}
\usepackage[rightcaption]{sidecap}
\definecolor{red}{rgb}{0.79509804490953684,0.20098039414733648,0.20686274673789687}
\definecolor{blue}{rgb}{0.27900615537575646,0.48814110222692564,0.65689543960433383}
\definecolor{green}{rgb}{0.35022877099759442,0.63449636544374854,0.34172819480299954}
\definecolor{purple}{rgb}{0.56810267475598009,0.34858709455412984,0.591251461777617}
\definecolor{orange}{rgb}{0.87539215686858873,0.50482698971149964,0.12774509808012069}
\definecolor{yellow}{rgb}{0.8939061906626995,0.88929258034948033,0.29840446252594988}
\definecolor{pink}{rgb}{0.90225298544939825,0.56472128699807567,0.74206845059114346}

\usepackage{array}
\newcolumntype{L}[1]{>{\raggedright\let\newline\\\arraybackslash\hspace{0pt}}m{#1}}
\newcolumntype{C}[1]{>{\centering\let\newline\\\arraybackslash\hspace{0pt}}m{#1}}
\newcolumntype{R}[1]{>{\raggedleft\let\newline\\\arraybackslash\hspace{0pt}}m{#1}}
\urldef{\mail}\path|{p.moeskops}@tue.nl|

\begin{document}

\mainmatter  

\title{Isointense infant brain MRI segmentation with a dilated convolutional neural network}

\titlerunning{Isointense infant brain MRI segmentation with a dilated CNN}
%
%
\author{Pim Moeskops \and Josien P.W. Pluim}

\authorrunning{P. Moeskops et al.}


\institute{Medical Image Analysis Group, Eindhoven University of Technology, The Netherlands}

%
%

\maketitle

\begin{abstract}
Quantitative analysis of brain MRI at the age of 6 months is difficult because of the limited contrast between white matter and gray matter. In this study, we use a dilated triplanar convolutional neural network in combination with a non-dilated 3D convolutional neural network for the segmentation of white matter, gray matter and cerebrospinal fluid in infant brain MR images, as provided by the MICCAI grand challenge on 6-month infant brain MRI segmentation.
\end{abstract}

\section{Introduction}
Brain MRI is important for quantitative analysis of neurodevelopment at all ages. A particularly difficult age in the context of brain MRI analysis is the age of 6 months, where the contrast between white matter and gray matter is limited because of the development of myelination \cite{Wang15,Zhan15}. 

Convolutional neural networks (CNNs) are an effective approach for infant brain MRI segmentation \cite{Zhan15,Moes16,Nie16}. A specific type of CNNs proposed for image segmentation are dilated CNNs, which can achieve a large receptive field using a limited number of trainable weights \cite{Yu16}. In medical imaging, dilated CNNs have been used for cardiac MRI segmentation \cite{Wolt16}.

In this study, we investigate using a dilated triplanar CNN in combination with a non-dilated 3D CNN for the segmentation of brain MRI of infants at an age of 6 months.

\section{Materials and Methods}
\subsection{Data}
The data used in this study is provided by the MICCAI grand challenge on 6-month infant brain MRI segmentation\footnote{This paper is a submission to the MICCAI grand challenge on 6-month infant brain MRI segmentation (\url{http://iseg2017.web.unc.edu/}).}. 10 images are provided as training data, 13 images are used to evaluate the method in the challenge. For all patients, T\textsubscript{1}- and T\textsubscript{2}-weighted images are provided.

To allow evaluation of the performance in initial experiments, 3 training images were randomly selected as validation set, the remaining 7 images are used to train the method. 

The images were preprocessed by the organisers of the challenge, this included skull stripping, intensity inhomogeneity correction and removal of the cerebellum and brain stem. As a reference standard, the images were manually segmented into white matter (WM), gray matter (GM) and cerebrospinal fluid (CSF).

\subsection{Method}
A CNN with triplanar dilated convolutions and 3D convolutions in four network branches is used to segment the images into WM, GM and CSF (Table \ref{tab:network}). 

The dilated branches consist of the same architecture as proposed by Yu et al. \cite{Yu16}, which uses layers of $3\times3$ kernels with increasing dilation factors. Input for the dilated branches is obtained from the axial, sagittal and coronal planes, as also proposed by Wolterink et al. \cite{Wolt16}. In contrast, our network uses three network branches that are combined in the last layer, instead of processing in three directions and averaging of the output probabilities. Moreover, in addition to these three dilated triplanar network branches, we include an additional 3D network with non-dilated 3D convolutions. All four network branches use 2-channel input from the T\textsubscript{1}- and the T\textsubscript{2}-weighted images. 

\begin{table}
\scriptsize
\begin{center}
\begin{tabular}[htb]{c c c c c c c c c c c}
\hline
\multicolumn{3}{|c}{\textbf{Axial}} & \multicolumn{3}{|c}{\textbf{Coronal}} & \multicolumn{3}{|c}{\textbf{Sagittal}} & \multicolumn{2}{|c|}{\textbf{3D}}\\\hline
\multicolumn{1}{|c}{Kernels} & \multicolumn{1}{|c}{Size} & \multicolumn{1}{|c}{Dilation} & \multicolumn{1}{|c}{Kernels} & \multicolumn{1}{|c}{Size} & \multicolumn{1}{|c}{Dilation} & \multicolumn{1}{|c}{Kernels} & \multicolumn{1}{|c}{Size} & \multicolumn{1}{|c}{Dilation} & \multicolumn{1}{|c}{Kernels} & \multicolumn{1}{|c|}{Size} \\\hline

\multicolumn{1}{|c}{32} & 3$\times$3 & 1 & 
\multicolumn{1}{|c}{32} & 3$\times$3 & 1 &  
\multicolumn{1}{|c}{32} & 3$\times$3 & 1 &
\multicolumn{1}{|c}{32} & \multicolumn{1}{c|}{3$\times$3$\times$3}\\

\multicolumn{1}{|c}{32} & 3$\times$3 & 1 &  
\multicolumn{1}{|c}{32} & 3$\times$3 & 1 &
\multicolumn{1}{|c}{32} & 3$\times$3 & 1 &
\multicolumn{1}{|c}{32} & \multicolumn{1}{c|}{3$\times$3$\times$3}\\

\multicolumn{1}{|c}{32} & 3$\times$3 & 2 & 
\multicolumn{1}{|c}{32} & 3$\times$3 & 2 & 
\multicolumn{1}{|c}{32} & 3$\times$3 & 2 &
\multicolumn{1}{|c}{32} & \multicolumn{1}{c|}{3$\times$3$\times$3}\\

\multicolumn{1}{|c}{32} & 3$\times$3 & 4 &   
\multicolumn{1}{|c}{32} & 3$\times$3 & 4 & 
\multicolumn{1}{|c}{32} & 3$\times$3 & 4 &
\multicolumn{1}{|c}{32} & \multicolumn{1}{c|}{3$\times$3$\times$3}\\

\multicolumn{1}{|c}{32} & 3$\times$3 & 8 &  
\multicolumn{1}{|c}{32} & 3$\times$3 & 8 &  
\multicolumn{1}{|c}{32} & 3$\times$3 & 8 &
\multicolumn{1}{|c}{32} & \multicolumn{1}{c|}{3$\times$3$\times$3}\\

\multicolumn{1}{|c}{32} & 3$\times$3 & 16 &
\multicolumn{1}{|c}{32} & 3$\times$3 & 16 & 
\multicolumn{1}{|c}{32} & 3$\times$3 & 16 &
\multicolumn{1}{|c}{32} & \multicolumn{1}{c|}{3$\times$3$\times$3}\\

\multicolumn{1}{|c}{32} & 3$\times$3 & 1 &
\multicolumn{1}{|c}{32} & 3$\times$3 & 1 & 
\multicolumn{1}{|c}{32} & 3$\times$3 & 1 &
\multicolumn{1}{|c}{32} & \multicolumn{1}{c|}{3$\times$3$\times$3}\\

\multicolumn{1}{|c}{} & & & \multicolumn{1}{|c}{} & & & \multicolumn{1}{|c}{} & & &\multicolumn{1}{|c}{32} & \multicolumn{1}{c|}{3$\times$3$\times$3} \\
\multicolumn{1}{|c}{} & & & \multicolumn{1}{|c}{} & & & \multicolumn{1}{|c}{} & & &\multicolumn{1}{|c}{32} & \multicolumn{1}{c|}{3$\times$3$\times$3} \\
\multicolumn{1}{|c}{} & & & \multicolumn{1}{|c}{} & & & \multicolumn{1}{|c}{} & & &\multicolumn{1}{|c}{32} & \multicolumn{1}{c|}{3$\times$3$\times$3} \\
\multicolumn{1}{|c}{} & & & \multicolumn{1}{|c}{} & & & \multicolumn{1}{|c}{} & & &\multicolumn{1}{|c}{32} & \multicolumn{1}{c|}{3$\times$3$\times$3} \\
\multicolumn{1}{|c}{} & & & \multicolumn{1}{|c}{} & & & \multicolumn{1}{|c}{} & & &\multicolumn{1}{|c}{32} & \multicolumn{1}{c|}{3$\times$3$\times$3} \\
\hline

\multicolumn{11}{|c|}{Concatenation of $4\times32=128$ feature maps}\\\hline

& & & & & & & & & \multicolumn{1}{|c}{3}& \multicolumn{1}{c|}{1$\times$1$\times$1}\\
\cline{10-11}\\
\end{tabular}
\end{center}
\caption{Combined 7-layer dilated triplanar CNN and 12-layer 3D CNN architecture. The dilated branches have receptive fields of $67\times67$ and the 3D branch has a receptive field of $25\times25\times25$. Because the background class is already defined by skull stripping, three output classes are used in the last layer (WM, GM and CSF).}
\label{tab:network}
\end{table}

Three versions of the network are evaluated in this study: (1) the dilated triplanar network with with shared weights between the three branches, (2) the dilated triplanar network with separate weights for each of the three branches, and (3) the dilated triplanar network  with separate weights combined with the 3D network as fourth branch.

Batch normalisation \cite{Ioff15} and ReLUs were used throughout. Dropout \cite{Sriv14} was used before the output layer. The network is trained with Adam \cite{King15} based on the cross-entropy loss, using mini-batches of 200 or 300 samples in 10 epochs of 50,000 random samples per class per training image.

The fully convolutional nature of all four branches allows arbitrarily sized inputs during testing. The method takes about 1 minute to segment a full image on a NVIDIA Titan X Pascal GPU.

\begin{figure}[htb]
\begin{center}
\begin{subfigure}{.24\textwidth}
\includegraphics[trim=3mm 6mm 3mm 9mm,clip,width=\textwidth]{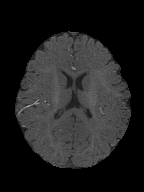}\\
\includegraphics[trim=3mm 6mm 3mm 9mm,clip,width=\textwidth]{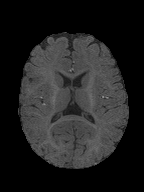}\\
\caption{T\textsubscript{1}-w image} 
\end{subfigure}
\begin{subfigure}{.24\textwidth}
\includegraphics[trim=3mm 6mm 3mm 9mm,clip,width=\textwidth]{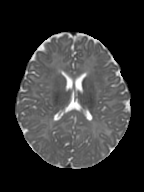}\\
\includegraphics[trim=3mm 6mm 3mm 9mm,clip,width=\textwidth]{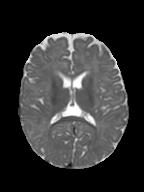}\\
\caption{T\textsubscript{2}-w image} 
\end{subfigure}
\begin{subfigure}{.24\textwidth}
\includegraphics[trim=3mm 6mm 3mm 9mm,clip,width=\textwidth]{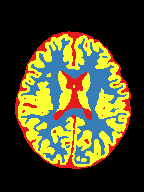}\\
\includegraphics[trim=3mm 6mm 3mm 9mm,clip,width=\textwidth]{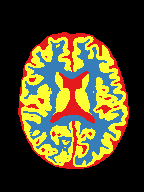}\\
\caption{Automatic} 
\end{subfigure}
\begin{subfigure}{.24\textwidth}
\includegraphics[trim=3mm 6mm 3mm 9mm,clip,width=\textwidth]{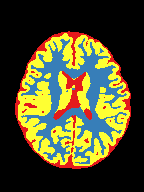}\\
\vspace{34.7mm}
\caption{Reference} 
\end{subfigure}
\end{center}
\caption{Example segmentations from the validation set (top), trained with 7 of the training images, and from the test set (bottom), trained with all 10 training images. From left to right: T\textsubscript{1}-weighted image (a), T\textsubscript{2}-weighted image (b), automatic segmenation (c), reference segmentation (d).} \label{fig:seg}
\end{figure}

\setlength{\tabcolsep}{1em}

\begin{table}
\scriptsize
\begin{center}
\begin{tabular}[htb]{l c c c}
\specialrule{1.5pt}{1pt}{1pt}
\multicolumn{4}{c}{\textbf{Triplanar network with shared weights}}\\
& WM & GM & CSF \\\hline
Validation image 1 & 0.859 & 0.874 & 0.914 \\
Validation image 2 & 0.818 & 0.795 & 0.905 \\
Validation image 3 & 0.891 & 0.870 & 0.897\\\hline
Average & 0.856 & 0.846 & 0.905 \\\specialrule{1.5pt}{1pt}{1pt}
\multicolumn{4}{c}{\textbf{Triplanar network with separate weights}}\\
& WM & GM & CSF \\\hline
Validation image 1 & 0.863 & 0.889 & 0.927 \\
Validation image 2 & 0.860 & 0.858 & 0.903 \\
Validation image 3 & 0.881 & 0.866 & 0.895 \\\hline
Average & 0.868 & 0.871 & 0.908 \\\specialrule{1.5pt}{1pt}{1pt}
\multicolumn{4}{c}{\textbf{Combined triplanar and 3D network}}\\
& WM & GM & CSF \\\hline
Validation image 1 & 0.885 & 0.909 & 0.948 \\
Validation image 2 & 0.830 & 0.821 & 0.921 \\
Validation image 3 & 0.905 & 0.900 & 0.925 \\\hline
Average & 0.874 & 0.877 & 0.932 \\
\specialrule{1.5pt}{1pt}{1pt}
\end{tabular}
\end{center}
\caption{Dice coefficients for the validation set, trained with 7 of the training images and evaluated on the remaining 3 images. From top to bottom: orthogonal triplanar dilated network with shared weights, orthogonal triplanar dilated network with separate weights, and orthogonal triplanar dilated network with separate weights and additional non-dilated 3D network.}
\label{tab:dice}
\end{table}

\newpage
\section{Results and Discussion}
Example segmentation results for the validation and test set are shown in Figure \ref{fig:seg}. The results on the validation set in terms of Dice coefficients are listed in Table \ref{tab:dice}. 

The results show accurate segmentation performance in 6-month isointense infant brain MR images. On the validation set, the performance of the triplanar network with separate weights was better than the performance of the network with shared weights. The best performance on the validation set was obtained using the network that combines the dilated triplanar network with a non-dilated 3D network (Table \ref{tab:network}). This combined network is therefore applied to the test set. The results on the test set will be available on the website of the MICCAI grand challenge on 6-month infant brain MRI segmentation\footnote{\url{http://iseg2017.web.unc.edu/results/}}.

\bibliographystyle{splncs03}
\bibliography{D:/literature/CAD}

\end{document}